\pdfoutput=1

\documentclass[11pt]{article}

\usepackage[final]{acl}

\usepackage{times}
\usepackage{latexsym}

\usepackage[T1]{fontenc}

\usepackage[utf8]{inputenc}

\usepackage{microtype}

\usepackage{inconsolata}

\usepackage[utf8]{inputenc}
\usepackage[english]{babel}
\usepackage{listings}  
\usepackage{url}
\usepackage{amsmath}
\usepackage{paralist}  
\usepackage{color}
\usepackage{theorem}
\usepackage{amsfonts}
\usepackage{amssymb}
\usepackage{amsmath}
\usepackage{bm}
\usepackage{dirtytalk}
\usepackage[export]{adjustbox}
\usepackage{graphicx}
\graphicspath{{Images/}}
\usepackage{caption}
\usepackage{subcaption}
\usepackage{makecell}
\usepackage{booktabs}
\usepackage{gb4e}
\noautomath

\title{A Hard Nut to Crack: Idiom Detection with Conversational Large Language Models}

\author{Francesca De Luca Fornaciari\textsuperscript{1}, Begoña Altuna\textsuperscript{2}, Itziar Gonzalez-Dios\textsuperscript{2}, Maite Melero\textsuperscript{1} \\ \textsuperscript{1}Barcelona Supercomputing Center (BSC) \\ \textsuperscript{2}HiTZ Center - Ixa, University of the Basque Country UPV/EHU \\  \texttt{[fdelucaf,maite.melero]@bsc.es}, \texttt{[begona.altuna,itziar.gonzalezd]@ehu.eus}
}

\begin{document}
\maketitle
\begin{abstract}
In this work, we explore idiomatic language processing with Large Language Models (LLMs). We introduce the Idiomatic language Test Suite IdioTS, a new dataset of difficult examples specifically designed by language experts to assess the capabilities of LLMs to process figurative language at sentence level. We propose a comprehensive evaluation methodology based on an idiom detection task, where LLMs are prompted with detecting an idiomatic expression in a given English sentence. We present a thorough automatic and manual evaluation of the results and an extensive error analysis.
\end{abstract}

\section{Introduction}
\label{sec:introduction}

The continuous improvements in LLM performance raise the hypothesis that their exposure to vast amounts of pre-training data may give them the capability to accurately process the meaning of natural language utterances.
We conducted a thorough analysis of the behaviour of three small-sized, instruction-tuned LLMs, tasked with figurative uses of language. The goal of this work is to provide a comprehensive evaluation methodology centred around a new test suite, IdioTS,\footnote{The resource is published under an open licence (CC BY-SA-NC 4.0) and can be accessed at this URL: https://ixa.si.ehu.es/node/14017} designed to assess the capabilities of LLMs to distinguish between figurative and literal meanings of Potentially Idiomatic Expressions (PIEs).
The adopted definition of PIE is the one provided by \citet{haagsma-etal-2020-magpie}: expressions that can have an idiomatic meaning, regardless of whether they actually have that meaning in a given context.

\section{Related work}
\label{related-work}

The question about to what extent LLMs can interpret non-literal phrases remains open \citep{jhamtani-etal-2021-investigating}. The creation of numerous figurative language datasets as a fundamental resource for evaluation underscores the importance of this issue in Natural Language Processing (NLP). To the best of our knowledge, these are some of the most significant existing datasets on figurative language.\\
The MAGPIE corpus, created by \citet{haagsma-etal-2020-magpie}, is a large sense-annotated corpus of PIEs created from a highly curated list of idioms. This dataset has been employed in numerous studies \citep{tan-jiang-2021-bert, madabushi2021astitchinlanguagemodels, dankers-etal-2022-transformer}, exploring figurative language processing from the most diverse perspectives.\\
The Fig-QA dataset was developed by \citet{liu-etal-2022-testing} to test the ability of LLMs to reason about figurative language. The findings of the conducted experiments underscored that LLMs still fall short of human performance, particularly in zero- or few-shot settings.\\
The IMPLI dataset \citep{stowe-etal-2022-impli} is a human annotated dataset consisting of paired sentences spanning idioms and metaphors, designed for natural language inference (NLI). The task consists in predicting whether the meaning of one text fragment (premise) entails another (hypothesis). Experiment findings indicate that, even when pre-training data includes figurative sentences, idiomatic language remains a challenge for pre-trained language models.\\
The ID10M multilingual dataset developed by \citet{tedeschi-etal-2022-id10m} was proposed as part of a complete framework for idiom identification in several languages.
The conducted experiments demonstrate that a model fine-tuned on this dataset is able to correctly predict the majority of idiomatic PIEs, but struggles with literal PIEs, tending to attribute them an idiomatic meaning.\\
The FLUTE dataset, introduced by \citet{chakrabarty2022flute}, is a dataset of textual explanations of figurative expressions. The results of the experiments conducted with models fine-tuned on FLUTE showed how such dataset can contribute to developing models that understand figurative language through textual explanations.

\section{Dataset creation process}
\label{sec:dataset-creation}

We introduce a new evaluation dataset specifically crafted for idiom detection in English. The rationale behind the creation of a new resource from scratch, rather than building on a pre-existing dataset, is grounded in the need to avoid data contamination by providing data that we can guarantee the assessed models have not seen before. 
In this section, we describe all the steps involved in the creation process of the Idiomatic language Test Suite IdioTS.

\subsection{Idioms list creation}
\label{sec:idioms-list-creation}

Drawing inspiration from \citet{haagsma-etal-2020-magpie}, we manually built a highly curated list of idioms extracted from diverse online platforms, such as \textit{Amigos Ingleses},\footnote{https://www.amigosingleses.com/} \textit{The idioms},\footnote{https://www.theidioms.com/} and \textit{EF English idioms}.\footnote{https://www.ef.com/wwen/english-resources/english-idioms/}
We selected the idioms with a view to producing a sufficiently comprehensive list in terms of diversity of syntactic structures. We included not only phrases with a completely fixed morpho-syntactic structure (``Nothing to write home about''), but also constructions with a high morpho-syntactic variability (``To blow your own trumpet'').  The idioms within the resource encompass verb-object constructions (``Hold your horses''), a wide range of structures with the verb ``to be'' followed by a prepositional phrase (``To be on the ball'', ``To be up your street''), adjective-noun combinations (``Cold turkey''), more or less complex prepositional phrases (``By the skin of your teeth'', ``Out of the blue''), binomial pairs consisting of two nouns linked by a conjunction (``Bits and bobs''), and appositional compounds (``Easy-peasy''), among others.
The idioms included in our list pertain to a colloquial text style and are frequent in spoken everyday language.

As a following step, we meticulously reviewed the idioms list to ensure a high degree of homogeneity. For syntactically flexible and semi-fixed expressions, adjustments were made by placing the main verb in the infinitive tense and in the active form. Personal pronouns and determiners were replaced with indefinite pronouns (e.g. ``It serves you right'' became ``To serve someone right''). Idioms with a fixed morpho-syntactic structure were preserved in their original form (e.g. ``Don't quote me'', ``Hold your horses''), as this is the sole form in which they appear in authentic usages.
The resulting database consists of 93 idioms, each associated with a unique alphanumeric identifier and the original source from which it was extracted.

\subsection{Idiomatic sentence crafting}
\label{sec:idiomatic-sentence-crafting}

Even though for the majority of the idioms an example sentence was provided in the original source, we decided to craft entirely new sentences in order to minimise the risk of data contamination.

As crowdsourcing has become increasingly popular for language resource development in NLP applications \citep{drutsa-etal-2021-crowdsourcing}, and is considered a valid method to outsource data generation by mitigating potential researcher bias, we organised a small-scale crowdsourcing on a voluntary basis.
To ensure the quality of the generated sentences, we established the essential requirements collaborators had to fulfil: native English speakers, predominantly of British origins, with a demonstrated high linguistic proficiency attaining at least a C1 level.

Collaborators were eight language professionals with a linguistic background (English teachers, linguists, translators, and NLP experts). 
They were provided with a spreadsheet containing just the idioms and an empty cell to fill with a sentence, without any additional context. They were instructed to select a few idioms of their choice and to craft a sentence per chosen idiom. They were asked to produce sentences representative of natural, spontaneous language use, provided it resonated authentically with their native speaker experience. An idiom with its corresponding sentence was included as an example.
Through this initiative we obtained the 164 idiomatic sentences corresponding to the positive class of our dataset.

\subsection{Distractor sentence crafting}
\label{sec:distractor-sentence-crafting}

At this point, the dataset needed to be augmented with instances of the negative class, i.e. plausible, grammatically and syntactically correct sentences containing a set of words that might belong to an idiomatic expression, but in fact are employed in a less common, literal way. These are meant to be the most challenging portion of our dataset. Whereas the interpretation of the meaning of distractor sentences would pose minimal difficulty for a human reader, our intuition was that a LLM would encounter issues with this particular type of sentences.
The complex task of generating this kind of sentences was undertaken internally to ensure both their quality and correctness, while also providing a subtle suggestion of idiomaticity.
We employed various approaches. Whenever possible, a new sentence was crafted by selecting the complete set of words composing the idiom and placing it unchanged in a different semantic context that, from a human perspective, unequivocally determined its literal meaning.
This happened, for instance, with the idiom idi07 ``Bob's your uncle'' (\ref{ex1}):
\begin{exe}
\ex \label{ex1} What a surprise! I didn't know Bob's your uncle.
\end{exe}

For other idioms, like idi25 ``It's like talking to a brick wall'', not the entire expression but only certain elements --- the verb ``to talk'' and the noun phrase ``brick wall'' --- were extracted and placed in a different context that changed their meaning to literal (\ref{ex2}):
\begin{exe}
\ex \label{ex2} Let's talk about how a brick wall can add charm and character to any space.
\end{exe}

In other cases, the applied strategy was to use some of the words composing the idiom with a different syntactic or even morphological role, like it happened for the idiom idi82 ``To make a living'' (\ref{ex3}):
\begin{exe}
\ex \label{ex3} I bought a new lamp and lots of plants to make our living room warmer and more cosy.
\end{exe}

One final employed method involved proposing an expression with a certain character overlapping and assonance with the idiom, for example ``speed and span'' and ``spick and span'' (\ref{ex4}):
\begin{exe}
\ex \label{ex4} It is difficult to measure the speed and span of the dissemination of the virus.
\end{exe}

\subsection{Sentence proofreading and final layout}
\label{sec:sentence-proofreading-and-final-layout}

As far as possible, efforts were made to avoid having more than one PIE in a single sentence. This strategy aimed to simplify the comprehension and execution of the task for the models as well as the collection and analysis of the model's responses for the researchers.

Additionally, a concerted effort was made to mitigate gender bias within our newly developed resource. Whenever possible, gender-specific terms were either eliminated or neutralised, a large number of sentences were reformulated adopting a gender neutral first person plural (``we''/``us''), second person singular or plural (``you''), or third person plural (``they''). Since the gender neutralisation is not always possible due to grammatical or syntactical constraints, meticulous attention was devoted to ensuring a representation of feminine and masculine gender terms as balanced as possible throughout the dataset.

Finally, each sentence was assigned a unique alphanumeric identifier containing information about the related idiom and a suffix indicating whether it is an idiomatic or a distractor sentence.

The final Idiomatic language Test Suite IdioTS is composed by a total of 250 sentences, 164 of which are idiomatic and 86 distractor sentences.

\section{Experiment definition}
\label{sec:experiment-definition}

Our experimental focus was pointed at evaluating the ability of the selected LLMs to detect an idiomatic expression in a given sentence.
This experiment  falls within the context of ``idiom detection'' and  involved a binary sentence classification task, being the two classes to predict ``idiomatic'' (positive class) and ``non-idiomatic'' (negative class). 
The goal was to assess whether LLMs are able to accurately capture the meaning of a PIE, distinguishing between figurative and literal meaning based on the formulation of the sentence.

Assuming that the pre-training data for these models contained the specific PIEs far more frequently with idiomatic than with literal meaning, the models may be inclined to attribute a figurative meaning to the expression based on probability distribution.


\subsection{Assessed LLMs}
\label{assessed-llms}

Ensuring a fair comparability among models is an unresolved challenge, due to the many internal aspects of a model that remain undisclosed. Nevertheless, for the scope of this study, we attempted to minimise differences, focusing on three LLMs that have the following characteristics in common: they have a transformer-based architecture and approximately 7 billion parameters in size, they are open source and fine-tuned for dialogue. The preference for open-source over proprietary models was motivated by transparency and reproducibility reasons, along with cost implications. The choice of the smallest model within a specific model family was motivated by the possibility to conduct experiments in a resource-efficient way, by using a local machine without a GPU. The choice of instruction fine-tuned, conversational models was based on the idea of simulating a real-world scenario where a user employs a chatbot application to solve a task or find an answer to a question.

In accordance with these considerations, we included the following models in our assessment:

\begin{itemize}
    \item Llama-2-7b-chat \citep{Touvron2023Llama2O}.
    \item Mistral-7b-Instruct \citep{Jiang2024MixtralOE}.
    \item Vicuna-7b \citep{zheng2023judging}.
\end{itemize}
 
Regarding configuration, we maintained default values for most hyper-parameters, such as top-k: 40 and top-p: 0.95, as we observed that altering these values in the development phase did not significantly impact the output. However, we had to extend the default token limit to 800 to accommodate the long prompt and the verbose model responses, and prevent errors related to exceeding the maximum token length.
We also set the temperature to 0 in order to make the model output deterministic and the experiment reproducible.

\subsection{Prompt engineering}
\label{sec:prompt-engineering}

At a broad level, the key of successful prompts lies in incorporating all necessary information while avoiding excessively complex instructions.
For our experiment, we employed the following question as the central component of the prompt:
\say{Is there an idiom in the sentence?}, followed by the sentence to analyse.

Conversational LLMs typically accept prompts structured in two parts: the system prompt, a generic instruction about the models behaviour in interactions, and the user prompt, containing the specific question or request.
In development, we accurately chose the optimal prompt structure for our experiment, which is exemplified in Appendix \ref{app:prompt}, Figure 1 and contains all the elements listed in the following lines.


\textit{Defining the persona}. This technique consists in assigning the model a specific role by including a short description in the prompt. In our case, we adopted this formulation: \say{You are a professional linguist specialising in figurative language}. Introducing the concept of ``figurative language'' we intended to guide the model to focus on this specific linguistic phenomenon. However, we acknowledge the potential risk of introducing some level of researcher bias.

\textit{Describing the task}. This was expressed through this wording: \say{Your task is to analyse English sentences that may contain an idiom, also known as an idiomatic expression}. To ensure accurate language identification, we specified the language name. Additionally, we employed two distinct forms to refer to idiomatic expressions, aiming to provide the most precise task description.

\textit{Zero-shot prompting}. We added no examples to the prompt. Through this approach we intended to test the model's ability to perform the task based on the task description alone.

\textit{Including a definition of ``idiom''}. Due to the lack of an unique agreed-upon definition of idiom, we saw the need to include a concise definition, in an effort to narrow down the potential variations in model outputs: \say{A phrase, expression,  or group of words that has a meaning different from the individual meanings of the words themselves, and employed to convey ideas in a non-literal or metaphorical manner}.
 
\textit{Requiring an answer in JSON format}. In order to mitigate the issue of overgeneration related to conversational LLMs, an explicit instruction was added to guide the model to provide an answer in a JSON format, specifying the fields and the information to include in each field of the JSON file. This approach forced the model to provide all and only the required information, structured in a way that facilitated the collection and analysis of the output. The wording for this instruction was the following: \say{The response should be in strict JSON format including four fields}, where we specified header and content for each field as follows:
 \begin{itemize}
 \itemsep0em
    \item `hasIdiom': Is there an idiom in the sentence? Give a true/false answer.
    \item `idiom': Should include which is the idiom contained in the sentence.
    \item `meaning': Should explain the meaning of the identified idiom.
    \item `explanation': Should include a concise elaboration.
\end{itemize}

 



From a technical point of view, we used the llama-cpp-python binding\footnote{https://github.com/abetlen/llama-cpp-python} that supports inference for many LLMs models and played a crucial role in converting the standardised prompt into a specific input format compatible with each of the models during the inference process. As an example, in Appendix \ref{app:prompt}, Figure 2 we show the input format generated for Llama2, where the tags delimiting system and user prompt were replaced with the standard ones accepted by this particular model.

\section{Findings}
\label{sec:findings}

We established two different levels of evaluation for the experiment. The first level consists of a completely automatic evaluation, whereas the second level is complemented with a thorough manual evaluation and error analysis.

\subsection{First level of evaluation}
\label{sec:first-level-of-evaluation}

At the first level, we employed the following automatic metrics to assess the capability of a model to detect an idiom in a given English sentence: Accuracy, Misclassification Rate (MR), Recall, Specificity, Precision, and Balanced Accuracy.

These metrics offer a general overview of the behaviour of the models and facilitate comparisons.

In Table \ref{tab:automatic-metrics-three-models-idiom-detection} we present the aggregated results.
At a broad level, all three models fall within the same range of results, as they show close scores in terms of Accuracy and Misclassification Rate.

\begin{table}[h!]
\centering
\resizebox{\columnwidth}{!}{%
\begin{tabular}{lrrr}
                                           & \multicolumn{1}{l}{\textbf{Llama2}} & \multicolumn{1}{l}{\textbf{Mistral}} & \multicolumn{1}{l}{\textbf{Vicuna}} \\ \hline \hline
                                           
\multicolumn{1}{l}{\textbf{Accuracy} $\uparrow$}    & 0.656                                & 0.660                                 & \textbf{0.676}                       \\
\multicolumn{1}{l}{\textbf{MR} $\downarrow$}          & 0.344                                & 0.340                                 & \textbf{0.324}                       \\
\multicolumn{1}{l}{\textbf{Recall} $\uparrow$} & \textbf{1.0}                         & 0.896                                 & 0.988                                \\
\multicolumn{1}{l}{\textbf{Specificity} $\uparrow$} & 0.0                                  & \textbf{0.209}                        & 0.081                                \\
\multicolumn{1}{l}{\textbf{Precision} $\uparrow$}   & 0.656                                & \textbf{0.680}                        & 0.672                                \\
\multicolumn{1}{l}{\textbf{Balanced Accuracy} $\uparrow$} & 0.5                                  & \textbf{0.553}                        & 0.535                                \\ \hline
\end{tabular}
}
\caption{Automatic metrics calculated for the three models. Numbers in bold indicate which model achieved the best result for each metric. For Misclassification Rate, lower values are indicative of better performance, as denoted by the downward arrow.}
  \label{tab:automatic-metrics-three-models-idiom-detection}
\end{table}

When we observe further metrics, such as Recall and Specificity, we immediately notice a particular behaviour for Llama2. The model shows a Recall of a hundred percent, meaning that it correctly classified all the idiomatic sentences, and a Specificity of 0.0, meaning that it did not correctly classify any of the distractor sentences. In fact, Llama2 only provided positive answers.
This behaviour is known as \textit{acquiescence} or \textit{agreement bias} and consists in the model trying to always provide an answer that is compliant or satisfies the user request. As demonstrated by our experiment, this can have counterproductive effects, leading the model to provide inaccurate responses.

Specificity, also known as True Negative Rate, is especially significant in our study, since it expresses the number of distractor sentences that were correctly classified. Given our initial assumption about distractor sentences being especially challenging, a high score for this metric reflects a good performance within the scope of the proposed task.
Mistral not only exhibits the best Specificity score, but also a considerable lead over the other models, clearly demonstrating its superiority in this specific aspect.
Furthermore, it achieves the best score for Precision, even though the difference compared to the other models is less pronounced.

Even though Vicuna obtained slightly better scores than Mistral and Llama2 in terms of Accuracy and MR, we can observe that Mistral strikes the best score in terms of \textbf{Balanced Accuracy}. In our scenario, where the positive class in the dataset is double the size of the negative class, Balanced Accuracy is a more robust metric, and it provides a more reliable measure of classification performance in the face of imbalanced data.

\paragraph{Analysis of misclassifications}
\label{sec:analysis-of-misclassifications}

All incorrect classifications for Llama2 are of the type false positive. Regarding Mistral and Vicuna, the two models share a similar distribution of misclassifications, being false positive the predominant type for both. This indicates that the most common behaviour pattern across models was incorrectly attributing idiomaticity to a sentence that is not idiomatic.
Conversely, both models exhibit fewer misclassifications of the false negative type, suggesting that they were generally effective in identifying the presence of an idiomatic expression in a sentence.

These observations align with findings from \citet{tedeschi-etal-2022-id10m}, and with our initial intuition that, given the pre-training data likely contains the given PIEs with idiomatic meaning more frequently than with a literal meaning, the models tend to classify these expressions as idiomatic rather than literal based on probability distribution.

\subsection{Second level of evaluation}
\label{sec:second-level-of-evaluation}

In our study, for each sentence classified as idiomatic, the models were asked to additionally specify the detected idiomatic expression.
We observed that in a certain number of cases the models, despite correctly classifying a sentence as idiomatic, did not detect the correct idiom and rather identified some other part of the sentence as idiomatic, such as a phrasal verb, a collocation, or a single word.
This observation underscores that general metrics are insufficient to conclusively demonstrate the capability of a LLM to detect an idiom in a sentence and motivated us to perform an additional verification step to validate the accuracy of true positive classifications.
We calculated \textbf{True Positive Consistency} as the proportion of true positive predictions where the correct idiomatic expression was accurately identified as well.
This additional score allowed us to validate whether the models response was grounded in the correct reason.

Table \ref{tab:true-positive-consistency} displays True Positive Consistency values for the three analysed models. Mistral exhibits the best score, achieving a True Positive Consistency of 0.905, followed by Vicuna, and lastly Llama2.

\begin{table}[h!]
\centering
\resizebox{\columnwidth}{!}{%
\begin{tabular}{lcccc}
 &
  \multicolumn{1}{l}{\textbf{\begin{tabular}[c]{@{}l@{}}Idiomatic\\ sentences\\ (positive class)\end{tabular}}} &
  \multicolumn{1}{l}{\textbf{\begin{tabular}[c]{@{}l@{}}True\\ positives\end{tabular}}} &
  \multicolumn{1}{l}{\textbf{\begin{tabular}[c]{@{}l@{}}True positives\\ with correct\\ reason\end{tabular}}} &
  \multicolumn{1}{l}{\textbf{\begin{tabular}[c]{@{}l@{}}True\\ Positive\\ Consistency$\uparrow$\end{tabular}}} \\ \hline
  \hline
\textbf{Llama2}  & 164 & 164 & 138 & 0.841          \\
\textbf{Mistral} & 164 & 147 & 133 & \textbf{0.905} \\
\textbf{Vicuna}  & 164 & 162 & 144 & 0.889          \\ \hline
\end{tabular}%
}
\caption{True Positive Consistency values per model.}
\label{tab:true-positive-consistency}
\end{table}

\subsubsection{Error analysis}
\label{sec:error-analysis}

By carefully examining  the responses in the `idiom' and `meaning' fields, we identified the elements that the models incorrectly detected as an idiom and upon which they based their classification of the sentence as idiomatic.
We identified recurring error patterns across the three models.

\paragraph{True positive wrong reason error types}
\label{true-positive-wrong-reason-error-types}

Regarding true positive with wrong reason, all of the three models in some cases identified a phrasal verb as an idiom (``to run off'', ``to look up'').

In other cases, the models detected an element in the sentence and reported an existing idiom --- containing this element --- which meaning is unrelated with the analysed sentence, like in the following example (\ref{ex5}) from Llama2:
\begin{exe}
\ex \label{ex5} Just cut to the chase! What did she say about us using her \textit{ironing} board?
\begin{xlist}
\exi{Idiom:} \textit{Ironing} out differences.
\exi{Meaning:} To resolve conflicts or disagreements.
\end{xlist}
\end{exe}

In some other cases, the models reported an existing idiom, that seemed completely unrelated with the given sentence. Regarding Vicuna, it seems that the model engaged in an inverse reasoning process by firstly extracting the meaning of the idiomatic expression from the sentence and subsequently identifying a different idiomatic expression that conveys a similar meaning.
\begin{exe}
\ex \label{ex6} That's the last straw! I won't let you push me around any further!
\begin{xlist}
\exi{Idiom:} To break the camel's back.
\exi{Meaning:} To reach a point where one can no longer tolerate or endure something.
\end{xlist}
\end{exe}
As we can observe in  (\ref{ex6}), the explanation provided by the model is consistent with the meaning of the original idiom ``To be the last straw'', as well as the identified idiom ``To break the camel's back''.

A last detected pattern for true positive with wrong reason was the models inventing a non-existing idiom by using a single word or a chunk of the examined sentence as in  (\ref{ex7}). With 23 occurrences, Llama2 was the model that showed this error type with the highest frequency:
\begin{exe}
\ex \label{ex7} I'm going to go out on a limb and say they \textit{used the old model} for this task.
\begin{xlist}
\exi{Idiom:} \textit{Used the old model}.
\exi{Meaning:} To use something that is no longer current or fashionable.
\end{xlist}
\end{exe}
We observed that the meaning explanations offered by the models appear plausible and coherent. This represents a potential concern, as it may lead to an over-reliance on model outputs, especially in situations where users lack the necessary language knowledge to verify the semantic content of the response.

\paragraph{False negative error types}
\label{sec:false-negative-error-types}

In our study, false negatives are idiomatic sentences that were incorrectly classified as non-idiomatic. This type of classifications were entirely absent for Llama2.
Regarding Vicuna, this happened in two instances, where the model failed to detect the idioms ``To make up your mind'' and ``To be crystal clear''. However, the model successfully detected these same idioms in the context of a different sentence.
Mistral classified 17 instances as false negatives. Despite the higher number of cases, the model did not exhibit a consistent behaviour. In some cases, it demonstrated a lack of knowledge of a specific idiom, failing to recognise it in different contexts. In other cases, its ability to detect a specific idiom varied as the sentence context changed.
Due to the lack of consistency, we were unable to identify any discernible patterns or offer an explanation for the behaviour of the models for this specific error type.

\paragraph{False positive error types}
\label{sec:false-positive-error-types}

Regarding instances where models wrongly attributed an idiomatic meaning to a literal sentence, we identified two main error types.
By the first type, the models wrongly classified a sentence as idiomatic and erroneously attributed an idiomatic meaning to some random element of the sentence, like in the following example (\ref{ex8}) from Mistral:
\begin{exe}
\ex \label{ex8} We need to cut and tighten this rope: it is too slack!
\begin{xlist}
\exi{Idiom:} Cut and tighten.
\exi{Meaning:} To shorten or make something more concise.
\end{xlist}
\end{exe}
We consider that this type of error might be, to some extent, a consequence of researcher bias: by explicitly asking the model whether a sentence contains an idiom may force the model to actively search for an idiom, altering the neutrality of its response. Therefore, we defer a detailed analysis of this type of errors to future work.

Conversely, we consider the second type, where models wrongly classified a sentence as idiomatic and erroneously attributed an idiomatic meaning to the PIE associated with the sentence, of significant interest for our study. In these cases, it seems plausible to assume that the models might have fallen into the intentional ``traps'' we set by incorporating distractor sentences into our dataset.

\begin{table}[h!]
\centering
\resizebox{\columnwidth}{!}{%
\begin{tabular}{lccc}
 &
  \multicolumn{1}{l}{\textbf{\begin{tabular}[c]{@{}l@{}}Distractor\\ sentences\\ (negative class)\end{tabular}}} &
  \multicolumn{1}{l}{\textbf{\begin{tabular}[c]{@{}l@{}}False positives\\ associated\\ PIE: total$\downarrow$\end{tabular}}} &
  \multicolumn{1}{l}{\textbf{\begin{tabular}[c]{@{}l@{}}False positives\\ associated\\ PIE: ratio$\downarrow$\end{tabular}}} \\ \hline \hline
\textbf{Llama2}  & 86 & 55 & 0.640 \\
\textbf{Mistral} & 86 & 53 & 0.616 \\
\textbf{Vicuna}  & 86 & \textbf{47} & \textbf{0.546} \\ \hline
\end{tabular}%
}
\caption{Number of false positives with idiomatic meaning attributed to the associated PIE over total distractor sentences per model.}
\label{tab:false-positives-correct-reason-three-models}
\end{table}

Table \ref{tab:false-positives-correct-reason-three-models} presents, for each model, the ratio of distractor sentences where the model attributed an idiomatic meaning to the associated PIE over the total number of distractor sentences (86) in the dataset.
As we can observe, the three examined models exhibit a comparable behaviour, with Vicuna showing the smallest number of errors of this type.

\section{Conclusions and future work}
\label{sec:conclusions-and-future}

The use of figurative language is a complex linguistic phenomenon that poses hard challenges for LLMs. Despite its critical role within numerous NLP tasks, it still remains a relatively under-explored area of investigation.

In this work we addressed the specific domain of idiomatic expressions in English as a special case of figurative language use. As a part of our contribution:
\begin{itemize}
    \item We introduced the new Idiomatic language Test Suite IdioTS, manually curated by language experts, and covering especially challenging idiomatic and literal uses of language.
    \item We proposed a comprehensive methodology for the assessment of the linguistic capabilities of LLMs in relation to idiomatic language.
    \item We conducted an idiom detection experiment focused on the assessment of the capabilities of small conversational LLMs to detect idioms within ambiguous English sentences.
    \item We conducted a thorough manual evaluation and error analysis and observed the main behaviour patterns of LLMs within this task.
\end{itemize}

The findings from our study indicate that when it comes to capturing the meaning of an ambiguous sentence, LLMs struggle to distinguish between literal and idiomatic uses of language. In line with the observations in the literature, a high acquiescence or agreement bias was observed: LLMs tend to force the identification of an idiom by assigning idiomatic meaning to an aleatory element in the sentence. Additionally, they offer coherent explanations to reinforce their inaccurate answers, which can be a cause for concern.

As future research directions, we intend to broaden our experiments by extending them to one- and few-shot scenarios, by exploring other prompting techniques focused on mitigating researcher bias and incorporating the possibility to interact with conversational models in multi-turn conversations.

Regarding the proposed IdioTS, we plan to explore several data augmentation techniques to generate additional idiomatic and distractor sentences. Additionally, a categorisation of distractor types could be incorporated to gain an understanding of which constructions are the most challenging for the models. Moreover, we intend to translate the sentences into other languages to create a multilingual dataset and open a path for MT experiments aimed to investigate possible correlations between idiom detection and translation.

At a broad level, exploring models with different architectures, sizes, and hyper-parameter configurations could provide valuable insights into how these models characteristics relate to the capabilities of LLMs to process natural language and could open avenues for targeted experimentation, such as specific fine-tuning strategies, aimed at enhancing the performance of LLMs across various natural language tasks.






\section*{Acknowledgements}

This work has been partially funded by i) DeepR3 (TED2021-130295B-C31) funded by MCIN/AEI/10.13039/501100011033 and European Union NextGeneration EU/PRTR. ii)   Ixa group A type research group (IT-1805-22) funded by the Basque Government, iii) DeepKnowledge (PID2021-127777OB-C21) project funded by MCIN/AEI/ 10.13039/501100011033 and by ERDF A way of making Europe and  iv)  AWARE Commonsense for a new generation of natural language understanding applications (TED2021-131617B-I00) funded by MCIN/AEI /10.13039/501100011033 by the European Union NextGenerationEU/ PRTR.

\bibliography{custom}

\onecolumn
\appendix
\section{Prompt formats used for the idiom detection experiment}
\label{app:prompt}

In Figure 1 we present the content of the prompt passed to all the assessed models.

\begin{center}
    \fbox{\parbox{0.9\linewidth}    {{
            ``role'': ``system'', \\
            ``content": \\
You are a professional linguist specialising in figurative language and your task 
is to analyse English sentences that may contain an idiom, 
also known as an idiomatic expression. This is a definition of idiom: `A phrase, expression, or group of words that has a meaning different from the individual meanings of the words themselves, and employed to convey ideas in a non-literal or metaphorical manner'. \\
The response should be in strict JSON format including four fields:\\
 `hasIdiom': Is there an idiom in the sentence? Give a true/false answer.\\
 `idiom': Should include which is the idiom contained in the sentence.\\
 `meaning': Should explain the meaning of the identified idiom.\\
 `explanation': Should include a concise elaboration.\\
        }
        {
            ``role'': ``user'',\\
            ``content'': sentence
        }}}
\end{center}
\begin{center}
{Figure 1: Prompt passed to all the assessed models.}
\end{center}

In Figure 2 we present the specific layout of the prompt generated by the llama-cpp-python binding for Llama2.

\begin{figure*}[h!]
  \centering
  \includegraphics[height=50mm, width=0.9\textwidth, frame]{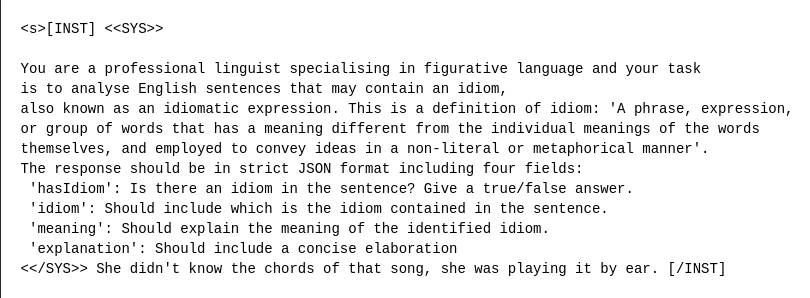}
  \label{fig:llama2-prompt-sentence-has-idiom}
\end{figure*}

\begin{center}
{Figure 2: Specific layout of the prompt generated by the llama-cpp-python binding for Llama2.}
\end{center}

\end{document}